\documentclass[journal]{IEEEtran}
\ifCLASSINFOpdf
\else
\fi

\usepackage{microtype}
\usepackage{graphicx}
\usepackage{subfigure}
\usepackage{booktabs} 

\usepackage{multirow}
\usepackage{url}            
\usepackage{bbm}
\usepackage{graphicx}
\graphicspath{{./figures/}}
\usepackage{amsmath}
\usepackage{color}

\begin{document}
%
\title{Single-Path Mobile AutoML: Efficient ConvNet Design and NAS Hyperparameter Optimization}

%
%

\author{Dimitrios~Stamoulis,~\IEEEmembership{Student Member,~IEEE,}
Ruizhou~Ding,~\IEEEmembership{Student Member,~IEEE,}
Di~Wang,~\IEEEmembership{Member,~IEEE,}
Dimitrios~Lymberopoulos,~\IEEEmembership{Member,~IEEE,}
Bodhi~Priyantha,~\IEEEmembership{Member,~IEEE,}
Jie~Liu,~\IEEEmembership{Fellow,~IEEE,}
and Diana~Marculescu,~\IEEEmembership{Fellow,~IEEE}
\thanks{D. Stamoulis, R. Ding, and D. Marculescu are with the
Department of ECE, Carnegie Mellon University, Pittsburgh, PA, USA 
(e-mail: dstamoul@andrew.cmu.edu).}
\thanks{D. Wang, D. Lymberopoulos, and B. Priyantha 
are with Microsoft, Redmond, WA, USA.}
\thanks{J. Liu is with the Harbin Institute of Technology, Harbin, China.}
\thanks{Manuscript received July 01, 2019; revised August xx, 20xx.}}

%
%

\markboth{}%
{}
%



\maketitle

\begin{abstract}
Can we reduce the search cost of Neural Architecture Search (NAS) from days down to only few hours? NAS methods automate the design of Convolutional Networks (ConvNets) under hardware constraints and they have emerged as key components of AutoML frameworks. However, the NAS problem remains challenging due to the combinatorially large design space and the significant search time (at least 200 GPU-hours). In this work, we alleviate the NAS search cost down to \textit{less than 3 hours}, while achieving state-of-the-art image classification results under mobile latency constraints. We propose a novel differentiable NAS formulation, namely \textit{Single-Path NAS}, that uses \textit{one} single-path over-parameterized ConvNet to encode all architectural decisions based on shared convolutional kernel parameters, hence drastically decreasing the search overhead. \textit{Single-Path NAS} achieves state-of-the-art top-1 ImageNet accuracy (75.62\%), hence outperforming existing mobile NAS methods in similar latency settings ($\sim$ 80ms). In particular, we enhance the accuracy-runtime trade-off in differentiable NAS by treating the Squeeze-and-Excitation path as a fully searchable operation with our novel \textit{single-path} encoding. Our method has an overall cost of only \textit{8 epochs} (24 TPU-hours), which is up to \textit{5,000$\times$ faster} compared to prior work. Moreover, we study how different NAS formulation choices affect the performance of the designed ConvNets. Furthermore, we exploit the efficiency of our method to answer an interesting question: instead of empirically tuning the hyperparameters of the NAS solver (as in prior work), can we automatically find the hyperparameter values that yield the desired accuracy-runtime trade-off (\textit{e.g.}, target runtime for different platforms)? We view our extensive experimental results as a valuable exploration for NAS-based cloud AutoML services, and we open-source our entire codebase at: \url{https://github.com/dstamoulis/single-path-nas}.
\end{abstract}


%
\IEEEpeerreviewmaketitle

\section{Introduction}

``\textit{Is it possible to automatically design the Convolutional Network (ConvNet) with highest classification accuracy that satisfies the inference latency constraints of a mobile phone? Can we have a push-button solution that automatically finds such design within only few hours?}'' ConvNets have been traditionally designed by human experts in a painstaking and expensive process. AutoML approaches, and Neural Architecture Search (NAS) methods in particular, present a promising path for alleviating the engineering costs that are intrinsic to the manual ConvNet design, by automating the tuning of DNN hyperparameters (\textit{e.g.}, the number of layers, the type of operations per layer, \textit{etc}). 

NAS approaches formulate the design of hardware-efficient ConvNets as a \textit{multi-objective hyperparameter optimization} problem~\cite{tan2018mnasnet}. In fact, we are witnessing a proliferation of novel AutoML approaches, with NAS formulations spanning many different optimization methodologies, such as Reinforcement learning~\cite{zoph2017learning}, evolutionary algorithms~\cite{real2018regularized}, and Bayesian optimization~\cite{kandasamy2018neural}. More importantly, NAS-based AutoML has drawn significant interest from industry, as demonstrated by the immense amount of computational resources used in NAS research~\cite{zoph2017learning,real2018regularized,bender2018understanding} and by the plethora of commercial cloud-based AutoML services and frameworks~\cite{botorch2019,golovin2017google,microsoft2019nni,google2019automl,google2018mnasnet}. Overall, AutoML is a research topic of paramount importance, since ``push-button'' solutions such as NAS frameworks are expected to significantly advance numerous deep learning applications, especially when designing ConvNets for computer vision tasks under the constraints of mobile devices~\cite{tan2018mnasnet}.

\begin{figure}[t!]
  \centering
  \includegraphics[width=0.98\columnwidth]{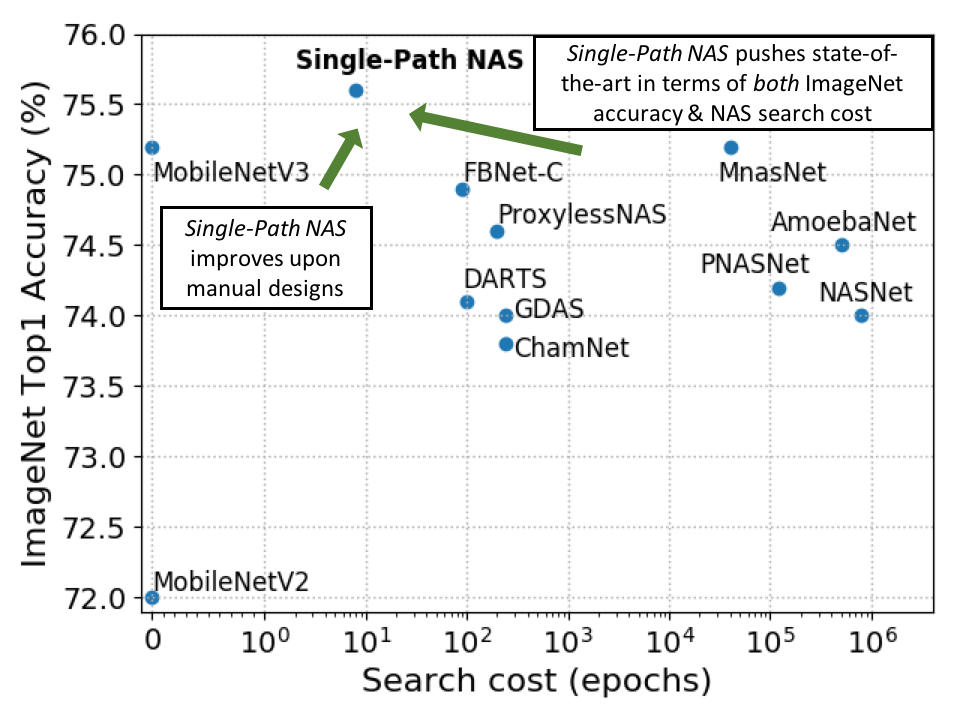}~
  \hspace*{80pt}\vspace{-82pt}
  \resizebox{.62\columnwidth}{!}{
	\begin{tabular}[b]{l|cc||l|cc}
		Method	& Top1 & Epochs & Method	& Top1 & Epochs  \\ \hline
        MobileNetV2~\cite{sandler2018mobilenetv2} & 72.0 & -- & MobileNetV3~\cite{howard2019searching}  & 75.2 & -- \\\hline
        NASNet-A~\cite{zoph2017learning}  & 74.0 &  800k  & Amoebanet-A~\cite{real2018regularized}  & 74.5 &  500k \\\hline
        PNASNet~\cite{liu2018progressive}  & 74.2 &  120k & DARTS~\cite{liu2018darts}  & 74.1 &  100 \\\hline
        GDAS-A1~\cite{tan2018mnasnet}  & 74.0 &  240  & MnasNet-A1~\cite{tan2018mnasnet}  & 75.2 &  40k \\\hline
        ChamNet-B~\cite{dai2018chamnet}   & 73.80 &  240  & FBNet-B~\cite{wu2018fbnet} & 74.9 & 90 \\\hline
        ProxylessNAS-R~\cite{cai2018proxylessnas} & 74.60 & 200 &  \textbf{Single-Path} NAS & \textbf{75.62} & \textbf{8} \\\hline
    \end{tabular}}
  \vspace{30pt}
  \caption{\textbf{Search Cost \textit{vs}. ImageNet Accuracy}: Our \textit{Single-Path NAS} outperforms Mobile NAS methods in both search cost and ImageNet top1 accuracy, while also improving upon manually-designed MobileNets~\cite{howard2019searching}. In particular, Single-Path NAS achieves new state-of-the-art 75.62\% top-1 accuracy compared to methods designing for similar latency setting ($\sim 80ms$). We report results from Mobile NAS and the ``Mobile setting'' of NAS literature (x-axis is shown in \texttt{symlog}-scale). Detailed discussion follows in Table~\ref{tab:imagenet-sota}.}
  \vspace{-10pt}
  \label{fig:key_idea}
\end{figure}

\begin{figure}[ht!]
  \centering
  \includegraphics[width=1.05\columnwidth]{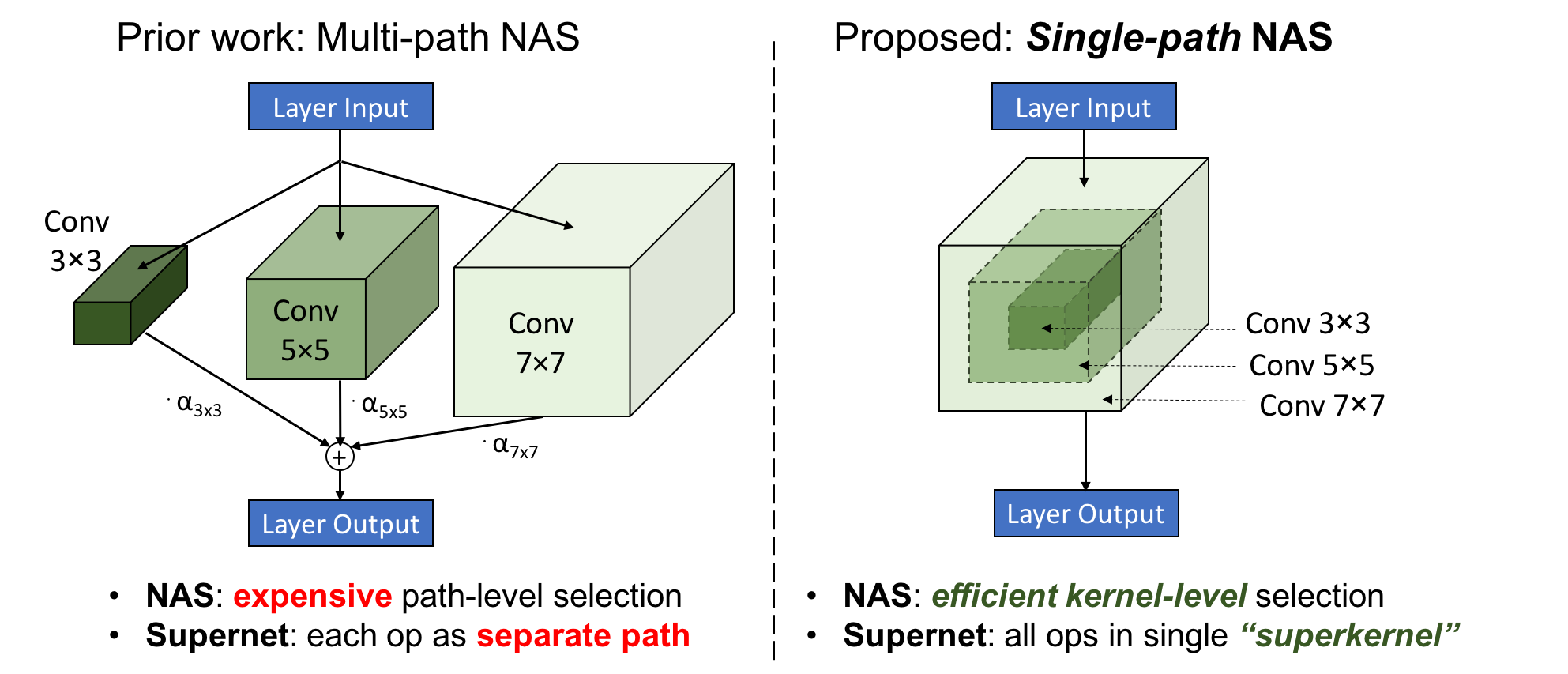}
  \vspace{-20pt}
  \caption{\textit{Single-Path NAS} directly 
  optimizes for the subset of convolution weights of 
  an over-parameterized superkernel in each ConvNet layer
  (right). Our \textbf{novel view} of the design space
  eliminates the need for maintaining separate paths for each candidate 
  operation, as in previous \textit{multi-path} approaches (left).}
  \vspace{-10pt}
  \label{fig:key_idea}
\end{figure}

Despite the recent breakthroughs, NAS remains an intrinsically costly  optimization problem due to the combinatorially large search space: \textit{e.g.}, for a mobile-efficient  ConvNet with 22 layers, choosing among five candidate operations yields $5^{22} \approx 2.3 \times 10^{15}$ possible ConvNet architectures. NAS literature has seen a shift towards one-shot differentiable formulations~\cite{liu2018darts,pham2018efficient,xie2018snas} which search over a supernet that encompasses all candidate architectures. Specifically, current NAS methods relax the combinatorial optimization problem of finding the optimal ConvNet  architecture to an operation/path selection problem: first, an over-parameterized, \textit{multi-path} supernet is constructed, where, for each layer, every candidate operation is added as a \textit{separate} trainable path, as illustrated in Figure~\ref{fig:key_idea} (left). Next, NAS formulations solve for the (distributions of) paths of the \textit{multi-path} supernet that yield the optimal architecture.

As expected, naively branching out all paths is intrinsically inefficient, since the number of trainable parameters that need to be maintained and updated during the search grows linearly with respect to the number of candidate operations per layer~\cite{bender2018understanding}. To tame the memory explosion introduced by the \textit{multi-path} supernet, current methods employ creative ``workaround'' solutions: \textit{e.g.}, searching on a proxy dataset~\cite{wu2018fbnet}, or employing a memory-wise scheme with only a subset of paths being updated during the search~\cite{cai2018proxylessnas}. However, these techniques 
remain considerably costly, with an overall computational demand of
hundreds of GPU-hours. 

In this paper, we propose \textit{Single-Path NAS}, a novel NAS method for designing
hardware-efficient ConvNets in \textbf{less than 3 hours}. Our \textbf{key insight} 
is illustrated in Figure~\ref{fig:key_idea} (right). We build upon the 
observation that different candidate convolutional operations in NAS 
can be viewed as subsets of a \textit{single superkernel}. Without having to 
choose among different paths/operations as in \textit{multi-path} methods, we instead 
solve the NAS problem as \textit{finding which subset of kernel weights to use 
in each ConvNet layer}. By sharing the convolutional kernel weights, 
we encode all candidate NAS operations into searchable superkernels (\textit{i.e.}, a single path) for each layer of the one-shot NAS supernet.  
Our contributions are as follows:

\textbf{1. Single-path differentiable NAS}: We propose a novel  \textit{single-path} encoding of the one-shot differentiable NAS problem. Moreover, while recent work investigates the use of Squeeze-and-Excitation~\cite{hu2018squeeze} (SE) as a binary NAS decision, we are first to treat the SE path as a fully searchable operation. To the best of our knowledge, this is the first single-path, differentiable NAS approach with SE paths, and our fully searchable treatment improves the accuracy-runtime trade-off compared to manually-tuned SE paths~\cite{howard2019searching}.

\textbf{2. State-of-the-art AutoML results}: \textit{Single-Path NAS} achieves $75.62\%$ top-1 accuracy on ImageNet with $\sim 80ms$ latency on a Pixel 1,  \textit{i.e.}, a $+0.4\%$ improvement over the current best hardware-aware NAS~\cite{tan2018mnasnet} and manually-designed~\cite{howard2019searching} ConvNets in similar latenct settings. The overall search cost is only 8 epochs, \textit{i.e.}, 2.45 hours on TPU-v3 (24 TPU-hours), up to \textbf{5,000$\times$ faster} compared to prior work. 

\textbf{3. NAS hyperparameter optimization}: To our knowledge, our work is the first to formulate the hyperparameter tuning of a differentiable NAS solver as a hyperparameter optimization problem itself, aiming to answer the question ``\textit{instead of empirically tuning, can we automatically find the trade-off hyperparameter in differentiable NAS given a target runtime?}''

\begin{figure*}[ht!]
  \centering
  \includegraphics[width=1.0\textwidth]{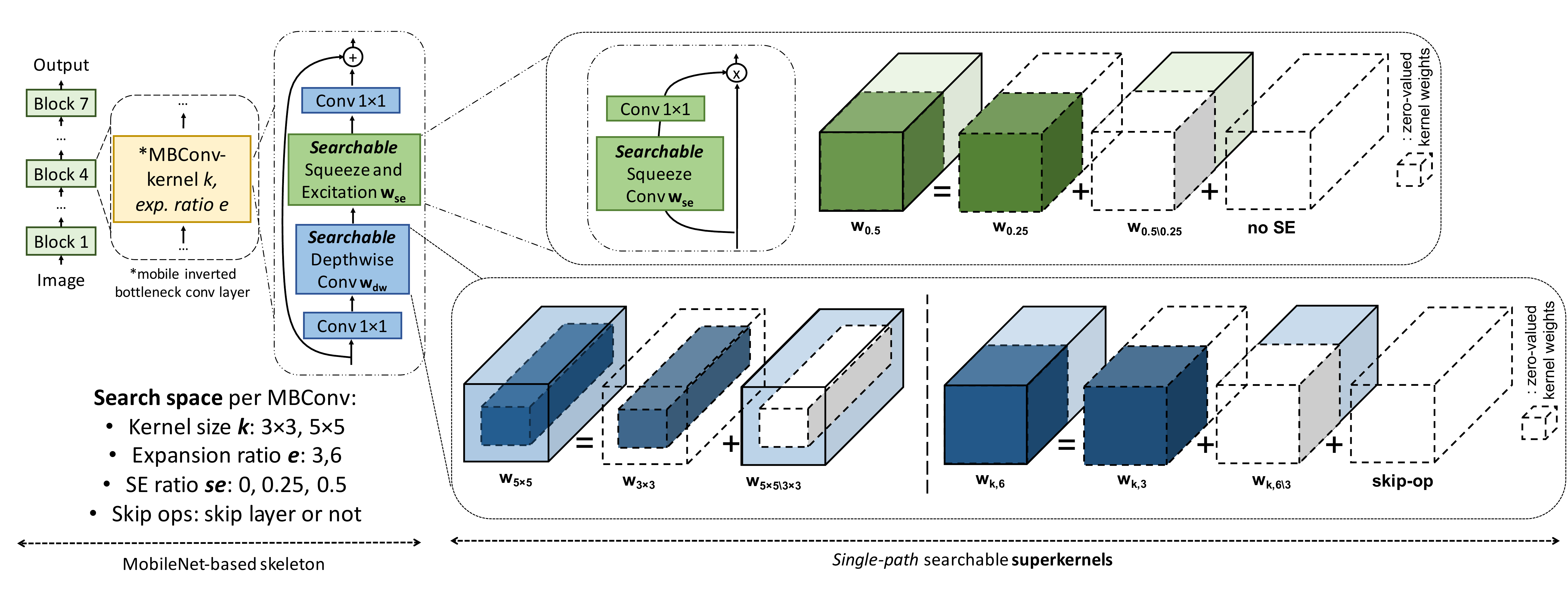}
  \vspace{-25pt}
  \caption{\textit{Single-Path} NAS builds upon \textit{hierarchical} MobileNetV2-like search spaces~\cite{tan2018mnasnet,stamoulis2019single} to identify the mobile inverted bottleneck convolution (MBConv) per layer (left). Our \textit{one-shot supernet} encapsulates all possible NAS architectures in the search space  without the need for appending each candidate operation as a separate path. \textit{Single-Path} NAS directly searches over the weights of two per-layer \textbf{searchable superkernels} that encode all MBConv types, \textit{i.e.}, the different kernel size (bottom, middle) and expansion ratio (bottom, right) values on the searchable \textit{depthwise}  superkernel, and the different Squeeze-and-Excitation~\cite{hu2018squeeze} (SE) ratios over the searchable \textit{squeeze} superkernel (top, right). That is, instead of treating the SE-path as a binary NAS decision (use it with fixed SE-ratio or not, as in~\cite{tan2018mnasnet}), we treat the SE path as a fully searchable operation with our \textit{single-path} encoding. We show that this search space enhancement further improves the accuracy-runtime trade-off.}
  \label{fig:design_space}
\end{figure*}

\section{Related Work}
\label{sec:rel}

While complex ConvNet designs
have unlocked unprecedented performance levels in computer vision tasks,
the accuracy improvement has come at the cost of higher computational 
complexity, making the deployment of state-of-the-art ConvNets to 
mobile devices challenging~\cite{stamoulis2018designing}. To this end, 
a significant body of prior work aims to co-optimize
for the inference latency of ConvNets. Earlier approaches focus on human expertise
to introduce hardware-efficient
operations~\cite{howard2017mobilenets,sandler2018mobilenetv2}.
Pruning~\cite{chin2018layer} and quantization~\cite{ding2017lightnn} methods 
share the same goal to improve the efficiency of ConvNets.

NAS methods aim to automate the design of ConvNets based 
on reinforcement learning (RL), evolutionary algorithms, or gradient-based 
formulations~\cite{liu2018darts,pham2018efficient,real2018regularized,zoph2016neural,zoph2017learning}.
Earlier approaches train an agent (\textit{e.g.}, RNN controller)
by sampling candidate architectures over a cell-based 
design space, where the same cell is repeated in all layers
and the focus is on searching the cell architecture~\cite{zoph2017learning}. Nonetheless, training the controller
over different architectures makes the search costly. An increasing number of recent methods motivate the need for alleviating the NAS search cost~\cite{dong2019searching}.

\textbf{Hardware-aware NAS}: Earlier NAS methods focused 
on maximizing accuracy under FLOPs constraints~\cite{xie2018snas,zhou2018resource}, 
but low FLOP count does not necessarily translate to hardware 
efficiency~\cite{dong2018dpp,stamoulis2018hyperpower}.
More recent methods incorporate hardware terms (\textit{e.g.}, runtime, power)
into cell-based NAS formulations~\cite{dong2018dpp,hsu2018monas}, but 
cell-based implementations are not hardware friendly~\cite{wu2018fbnet}.
Breaking away from cell-based assumptions in the search space encoding, Mnasnet 
searches over a generalized MobileNetV2-based design space~\cite{tan2018mnasnet}.

Recent NAS literature has seen a shift towards one-shot 
NAS formulations~\cite{liu2018darts,pham2018efficient,xie2018snas}. 
Differentiable NAS in particular has gained increased popularity and 
has achieved state-of-the-art results~\cite{brock2017smash}. One-shot-based 
methods use an over-parameterized super-model network, where, for each layer, every 
candidate operation is added as a separate trainable path.
Nonetheless, \textit{multi-path} 
search spaces have an intrinsic limitation: the number of trainable 
parameters that need to be maintained and updated with gradients 
during the search grows linearly with respect to the number of different
convolutional operations per layer, resulting in memory 
explosion~\cite{bender2018understanding,cai2018proxylessnas}.

To this end, state-of-the-art approaches employ different ``workaround'' solutions.
FBNet~\cite{wu2018fbnet} searches on a ``proxy'' dataset (\textit{i.e.},
subset of the ImageNet dataset). Despite the decreased search cost thanks 
to the reduced number of training images, these approaches do not address the fact 
that the entire supermodel needs to be maintained in memory during search, 
hence the efficiency is limited due to inevitable use of smaller batch sizes. 
ProxylessNAS~\cite{cai2018proxylessnas} employs a memory-wise 
scheme, where only a set of paths is updated during search. However, such implementation improvements do not address a second key suboptimality of one-shot approaches, 
\textit{i.e.}, the fact that separate gradient steps are needed to update the 
weights and the architectural decisions interchangeably~\cite{liu2018darts}. Although the number of trainable parameters in terms of memory cost is kept to the same level at any step, the way that \textit{multi-path}-based methods traverse the design space remains inefficient. 

While concurrent methods consider relaxed convolution
formulations with insight similar to our work~\cite{shin2018differentiable,hundt2019sharpdarts,guo2019single,stamoulis2019single}, 
they either use design spaces and objectives that have been shown to be 
hardware inefficient (\textit{e.g.}, cell-based space, FLOP count), or 
they optimize over a subset of our design space. In our work, we optimize over multiple searchable kernels per layer and we simultaneously search across several NAS decisions, \textit{i.e.}, kernel sizes, channels dimensions, expansion ratio, or Squeeze-and-Excitation~\cite{hu2018squeeze} ratio dimensions.

\textbf{Searching for Squeeze-and-Excitation~\cite{hu2018squeeze}}: Recently, MobileNetV3 explored various design choices on top of the MobileNetV2 backbone, showing that augmenting the mobile inverted bottleneck convolution (MBConv) layers with a Squeeze-and-Excitation~\cite{hu2018squeeze} (SE) path can improve the overall accuracy~\cite{howard2019searching}. Recent RL-based mobile NAS has adapted this finding by adding the SE path into their search space~\cite{tan2018mnasnet}, but by limiting however their exploration to a binary decision of using SE or not. Instead, in our work we are the \textit{first} to treat the SE path as fully searchable (\textit{i.e.}, searching over various SE ratios), with a novel outcome. As discussed in our results section, larger SE ratios further improve the overall performance by yielding a better DNN accuracy-trade-off. Our AutoML-designed DNN achieves a new state-of-the-art ImageNet accuracy compared to methods designing for similar latency settings ($\sim 80ms$).

\section{Proposed Method: \textit{Single-Path} NAS}
\label{sec:sp-methodology}

In this Section, we present our proposed 
method. First, we discuss our novel \textit{single-path} view 
(Subsection~\ref{subsec:view}) of the search space. Next, 
we encode the NAS problem as finding the 
subset of convolution weights over the \textit{over-parameterized} 
superkernel (Subsection~\ref{subsec:single-path-kernel}), and we discuss 
how it compares to existing \textit{multi-path}-based NAS 
(Subsection~\ref{subsec:comparison-vs-multi}). Last, we formulate 
the hardware-aware NAS objective function, where we incorporate 
an accurate inference latency model of ConvNets executing on the
Pixel~1 smartphone (Subsection~\ref{subsec:hw-loss}).

\subsection{Mobile ConvNets Search Space: A Novel View}
\label{subsec:view}

\textbf{Search Space}: As illustrated in Figure~\ref{fig:design_space} (left), our method builds upon a fixed ``backbone''~\cite{cai2018proxylessnas} which follows the MobileNetV2 design~\cite{sandler2018mobilenetv2} and which has been successfully considered by other differentiable NAS approaches~\cite{stamoulis2019single}. Specifically, in this macro-architecture, except for the head and stem layers, all ConvNet layers are grouped into blocks based on their filter sizes. The filter numbers per block follow the values in~\cite{wu2018fbnet}, \textit{i.e.}, we use seven blocks with up to four layers each. Each layer of these blocks is a mobile inverted bottleneck convolution MBConv~\cite{sandler2018mobilenetv2} micro-architecture. In particular, an MBConv layer consists of a point-wise ($1\times 1$) convolution, a $k\times k$ depthwise convolution, a Squeeze-and-Excitation (SE) block~\cite{hu2018squeeze}, and a linear $1\times 1$ convolution. Unless the layer has a stride value of two, a skip path is introduced to provide a residual connection from input to output. The goal of NAS is to automatically identify the type of each MBConv layer in the ConvNet design. 

Our search space consists of 13 candidate layer types, with the layer-wise choices listed in Figure~\ref{fig:design_space}. In particular, each MBConv layer is parameterized by the following choices: (i) the kernel size of the depthwise convolution $k \times k$, (ii) the expansion ratio $e$, \textit{i.e.}, the ratio between the output and input of the first $1\times 1$ convolution, and (iii) the Squeeze-and-Excitation~\cite{hu2018squeeze} (SE) ratio $se$, \textit{i.e.}, the ratio between the number of channels in the intermediate convolution and the input of the SE path. It is worth observing that, unlike prior NAS work, in our search space we treat the SE-path as fully searchable (\textit{i.e.}, searching over various SE ratios). Furthermore, NAS considers a special \textit{skip-op} ``layer'', which ``zeroes-out'' the kernel and feeds the input directly to the output, \textit{i.e.}, the entire layer is dropped. This NAS choice effectively corresponds to reducing the depth of the network. Based on this parameterization, we denote each MBConv as MBConv-$k\times k$-$e$-$se$.

\textbf{Novel view of design space}: 
We build upon the \textit{key observation} that different candidate convolutional operations in NAS can be viewed as subsets of the weights of over-parameterized \textit{superkernels}. This observation allows us to view the NAS combinatorial problem as \textit{finding which subset of kernel weights to use in each MBConv layer}, while sharing the kernel parameters across different MBConv architectural options. As shown in Figure~\ref{fig:design_space}, we encode all candidate NAS operations to two searchable superkernels (\textit{i.e.}, a \textit{single path}), for each layer of the one-shot NAS supernet.

\subsection{Proposed Methodology: Single-Path NAS formulation}
\label{subsec:single-path-kernel}

\textbf{Kernel size}:  
To simplify notation and without loss of generality, we show the case of choosing between a 
$3 \times 3$ or a $5 \times 5$ kernel for an MBConv layer. Let us denote the weights of the two candidate kernels as $\textbf{w}_{3 \times 3}$ and $\textbf{w}_{5 \times 5}$, respectively. 
As shown in Figure~\ref{fig:design_space} (bottom), we observe that 
the weights of the $3 \times 3$ kernel can be viewed as 
the \textit{inner} core of the weights of the $5 \times 5$ kernel, 
while ``zeroing'' out the weights of the ``\textit{outer}'' shell.
We denote this (\textit{outer}) subset of weights (that does not contribute 
to output of the $3 \times 3$ kernel but only to the $5 \times 5$ kernel), 
as $ \textbf{w}_{5 \times 5 \setminus 3 \times 3}$.
Hence, the NAS architectural choice of using 
the $5 \times 5$ convolution corresponds to using both 
the \textit{inner} $\textbf{w}_{3 \times 3}$ weights and the \textit{outer} shell, 
\textit{i.e.}, $\textbf{w}_{5 \times 5} = \textbf{w}_{3 \times 3} + 
\textbf{w}_{5 \times 5 \setminus 3 \times 3}$. 

We can therefore encode the NAS decision directly into the 
superkernel of an MBConv layer as a function of kernel weights
as follows:
\begin{equation}
    \label{eq:idea}
    \textbf{w}_{k} =  \textbf{w}_{3 \times 3} + \mathbbm{1}(\text{use~} 5 \times 5)  
    \cdot \textbf{w}_{5 \times 5 \setminus 3 \times 3}
\end{equation}
where $\mathbbm{1}(\cdot)$ is the indicator function that encodes the 
architectural NAS choice, \textit{i.e.}, if $\mathbbm{1}(\cdot) = 1$ then 
$\textbf{w}_{k} = \textbf{w}_{3 \times 3} + \textbf{w}_{5 \times 5 \setminus 3 \times 3} = 
\textbf{w}_{5 \times 5} $, else $\mathbbm{1}(\cdot) = 0$ then  
$\textbf{w}_{k} = \textbf{w}_{3 \times 3} $.

\textbf{Trainable encoding}: While the indicator function
encodes the NAS decision, a critical choice is how to formulate the condition over 
which the $\mathbbm{1}(\cdot)$ is evaluated. Our intuition is that, for an 
indicator function that represents whether to use the subset of weights,
its condition should be \textit{directly a function of the subset's weights}. 
Thus, our goal is to define an ``importance'' signal of the subset
weights that intrinsically captures their contribution to the overall ConvNet
loss. We draw inspiration from weight-based conditions that have been 
successfully used for quantization-related 
decisions~\cite{ding2019flightnns,choi2018pact} and we use 
the \textit{group Lasso term}. Specifically, for the indicator related to
the $\textbf{w}_{5 \times 5 \setminus 3 \times 3}$ ``outer shell'' decision, 
we write condition: 
\begin{equation}
    \label{eq:proposed-form-2}
    \textbf{w}_{k} = \textbf{w}_{3 \times 3} + \mathbbm{1}(\left\Vert 
    \textbf{w}_{5 \times 5 \setminus 3 \times 3} \right\Vert^2 > t_{k=5}) 
    \cdot \textbf{w}_{5 \times 5 \setminus 3 \times 3}
\end{equation}
where $t_{k=5}$ is a latent variable that controls the decision (\textit{e.g.},
a threshold value) of selecting kernel $5 \times 5$. The threshold will be compared to 
the Lasso term to determine if the \textit{outer}
$\textbf{w}_{5 \times 5 \setminus 3 \times 3 }$ weights are used to the overall convolution.
It is important to notice that, instead of picking the thresholds (\textit{e.g.}, $t_{k=5}$) 
by hand, we seamlessly 
treat them as trainable parameters to learn via gradient descent. 
To compute the gradients for thresholds, we relax the indicator 
function $g(x,t) = \mathbbm{1}(x>t)$ to a 
sigmoid function, $\sigma(\cdot)$, when computing gradients, \textit{i.e.}, 
$\hat{g}(x,t) = \sigma(x>t)$. 

\textbf{Expansion ratio and skip-op}: Since the result of the 
kernel-based NAS decision $\textbf{w}_{k}$ (Equation~\ref{eq:proposed-form-2}) is a 
convolution kernel itself, we can in turn apply our formulation to also encode 
NAS decisions for the expansion ratio of the $\textbf{w}_{k}$ kernel.
As illustrated in Figure~\ref{fig:design_space} (bottom, right), the channels of the 
depthwise convolution in an MBConv-$k\times k$-$3$ layer with expansion ratio $e=3$ 
can be viewed as using one half of the channels of an 
MBConv-$k\times k$-$6$ layer with expansion ratio $e=6$, while ``zeroing'' 
out the second half of channels $\{\textbf{w}_{k, 6 \setminus 3}\}$.
Finally, by ``zeroing'' out the first half of the output filters as well, 
the entire superkernel contributes nothing if added to the 
residual connection of the MBConv layer: \textit{i.e.}, by deciding if $e=3$,
we can encode the NAS decision of using, or not, only the ``skip-op'' path. For both decisions over the searchable kernel of the depthwise convolution, we write:
\begin{equation}
    \begin{split}
    \label{eq:proposed-form-3}
    \textbf{w}_{dw} = \mathbbm{1}(\left\Vert 
    \textbf{w}_{k, 3} \right\Vert^2 > & t_{e=3}) \cdot 
    ( \textbf{w}_{k, 3} + \\ & \mathbbm{1}(\left\Vert 
    \textbf{w}_{k, 6 \setminus 3} \right\Vert^2 > t_{e=6}) 
    \cdot \textbf{w}_{k, 6 \setminus 3} )
    \end{split}
\end{equation}

\textbf{SE ratio}: Next, we extend the superkernel-based definition to encode the Squeeze-and-Excitation~\cite{hu2018squeeze} (SE) ratio $se$ decision. In particular, we observe that choosing SE ratio (Equation~\ref{eq:proposed-form-3}) effectively means to choose the number of channels of the \textit{squeeze} convolution stage in the SE path. Hence, as shown in Figure~\ref{fig:design_space} (top, right), we replace the convolution kernel of the \textit{squeeze} convolution with a searchable superkernel, where the largest number of channels corresponds to the largest candidate $se$ value, \textit{i.e.}, $se=0.5$. By following an intuition similar to Equation~\ref{eq:proposed-form-3}, we observe that ``zero-ing out'' the second half of the  \textit{squeeze} convolution corresponds to using $se=0.25$, while ``zero-ing out'' the entire kernel corresponds to not using a SE path ($se=0$). We therefore write:
\begin{equation}
    \begin{split}
    \label{eq:proposed-form-4}
    \textbf{w}_{se} & = \mathbbm{1}(\left\Vert 
    \textbf{w}_{0.25} \right\Vert^2 > t_{se=0.25}) \\ & \cdot 
    ( \textbf{w}_{0.25} + \mathbbm{1}(\left\Vert 
    \textbf{w}_{0.5 \setminus 0.25} \right\Vert^2 > t_{se=0.5}) 
    \cdot \textbf{w}_{0.5 \setminus 0.25} )
    \end{split}
\end{equation}

\textbf{Searchable MBConvs}: Each MBConv uses $1\times 1$ convolutions for the point-wise and linear stages, while the kernel-size decisions affect only the $k\times k$ depthwise convolution (Figure~\ref{fig:design_space}). Thus, we use our \textit{searchable} depthwise kernel $\textbf{w}_{dw}$ at this middle stage. In terms of number of channels, the depthwise kernel depends on the point-wise $1\times 1$ output, which allows us to encode the expansion ratio $e$ into $\textbf{w}_{dw}$ as well. That is, we set the point-wise $1\times 1$ output to the maximum candidate expansion ratio, we instead solve for which of them not to ``zero'' out at the depthwise stage. In other words, we also encode the NAS decision for the expansion ratio at $\textbf{w}_{dw}$. Similarly, we can encode the SE ratio $se$ by deciding which the channels of the $1\times 1$ \textit{squeeze} convolution to ``zero'' out. To this end, we can simply replace the \textit{squeeze} kernel with the \textit{searchable} kernel $\textbf{w}_{se}$ to directly search for the SE-ratio across the SE path (Figure~\ref{fig:design_space}, top right). 

Overall, our \textit{single-path} formulation can sufficiently capture any MBConv type (\textit{e.g.}, MBConv-$3\times 3$-$6$-$0.25$, MBConv-$5\times 5$-$3$-$0.5$, \textit{etc.}) in the design space (Figure~\ref{fig:design_space}). For  input $\textbf{x}$, the output of the $i$-th MBConv layer of the network is: 
\begin{equation}
    \label{eq:effective-kernel-output}
    o^i (\textbf{x}) = \text{conv}(\textbf{x}, \textbf{w}^{i} | t_{k=5}^{i}, t_{e=6}^{i}, t_{e=3}^{i}, t_{se=0.5}^{i}, t_{se=0.25}^{i})
\end{equation}

\subsection{Single-Path vs. Existing Multi-Path Assumptions}
\label{subsec:comparison-vs-multi}

We briefly illustrate how our \textit{single-path} formulation compares to multi-path 
NAS approaches. In existing methods~\cite{cai2018proxylessnas,liu2018darts,wu2018fbnet},
the output of each layer $i$ is a (weighted) sum defined over the output of $N$ 
different paths, where each path $j$ corresponds to a different candidate 
kernel $\textbf{w}^{i,j}_{k \times k, e}$. The weight of each path 
$\alpha^{i,j}$ corresponds to the probability that this path 
is selected over the parallel paths:
\begin{equation}
    \begin{split}
    \label{eq:comparison}
    o^{i}&_{multi-path}(\textbf{x}) = \sum_{j=1}^N \alpha^{i,j} \cdot o^{i,j}(\textbf{x}) \\ = &
    \alpha^{i,0} \cdot \text{conv}(\textbf{x}, \textbf{w}^{i,0}_{3 \times 3}) + \dots + \alpha^{i,N} 
    \cdot \text{conv}(\textbf{x}, \textbf{w}^{i,N}_{5 \times 5})
    \end{split}
\end{equation}
It is easy to see how our novel \textit{single-path} view is advantageous, 
since the output of the convolution at layer $i$ of our search space 
is \textit{directly a function of the weights of our single over-parameterized kernel} 
(Equation~\ref{eq:effective-kernel-output}): 
\begin{equation}
    \begin{split}
    \label{eq:single-path-conv}
    o^{i}&_{single-path} (\textbf{x}) = o^i (\textbf{x}) \\ & =  \text{conv}(\textbf{x}, \textbf{w}^{i} | 
    t_{k=5}^{i}, t_{e=6}^{i}, t_{e=3}^{i}, t_{se=0.5}^{i}, t_{se=0.25}^{i})
    \end{split}
\end{equation}

Multi-path NAS methods solve for the optimal architecture parameters 
$\alpha$ (path weights), such that the weights 
$w_{\alpha}$ of the corresponding $\alpha$-architecture have minimal 
loss $\mathcal{L} (\alpha, w_{\alpha})$:
\begin{equation}
    \label{eq:bilevel}
    \underset{\alpha}{\text{min }} \underset{w_{\alpha}}{\text{min }} \mathcal{L}(\alpha, w_{\alpha})
\end{equation}
However, solving Equation~\ref{eq:bilevel} gives rise to a challenging \textit{bi-level} 
optimization problem~\cite{liu2018darts}. Existing methods interchangeably 
update the $\alpha$'s while freezing the $w$'s and vice versa, leading to more gradient steps.

In contrast, with our \textit{single-path} formulation, the overall network loss
is directly a function of the superkernel weights, where the learnable
kernel- and expansion ratio-related threshold variables, $\textbf{t}_k$
and $\textbf{t}_e$, are directly derived as a function (norm) of
the kernel weights $\textbf{w}$. Consequently, \textit{Single-Path NAS} formulates 
the NAS problem as solving \textit{directly over the weight kernels $\textbf{w}$ 
of a single-path, compact neural network}. Formally, the NAS problem becomes:
\begin{equation}
    \label{eq:sp-nas}
    \underset{\textbf{w}}{\text{min }} \mathcal{L}(\textbf{w} | \textbf{t}_{k}, \textbf{t}_{e}, \textbf{t}_{se})
\end{equation}

\textbf{Efficiency of \textit{Single-Path NAS}}: Unlike the bi-level optimization problem in prior work, solving our NAS formulation in Equation~\ref{eq:sp-nas} is as expensive as training the weights of a single-path, \textit{branchless}, compact neural network with vanilla gradient descent. Therefore, our formulation eliminates the need  for separate gradient steps between the ConvNet weights and the NAS parameters. Moreover, the reduction of the trainable  parameters $\textbf{w}$ per se, further leads to a drastic reduction of the search cost down to \textbf{just a few epochs}, as our experimental results show later in Section~\ref{sec:results}.

\subsection{Hardware-Aware NAS with Differentiable Runtime Loss}
\label{subsec:hw-loss}

To design hardware-efficient ConvNets, the differentiable objective in Equation~\ref{eq:sp-nas}
should reflect both the accuracy of the searched architecture and its inference latency
on the target hardware. Hence, we use a latency-aware 
formulation~\cite{wu2018fbnet}:
\begin{equation}
    \label{eq:loss}
    \mathcal{L}(\textbf{w} | \textbf{t}_{k}, \textbf{t}_{e}, \textbf{t}_{se} ) = 
    CE( \textbf{w} | \textbf{t}_{k}, \textbf{t}_{e}, \textbf{t}_{se} ) + \lambda \cdot 
    \text{log}(R(\textbf{w} | \textbf{t}_{k}, \textbf{t}_{e}, \textbf{t}_{se}))
\end{equation}
The first term $CE$ corresponds to the cross-entropy loss of the
single-path model. The hardware-related term $R$ is the 
runtime in milliseconds ($ms$) of the searched NAS model on the target 
mobile platform. Finally, the coefficient $\lambda$ modulates
the trade-off between cross-entropy and runtime.

To preserve the differentiability of the objective, another critical choice is the formulation of
the latency term $R$. Prior art has showed that the total network latency of a 
mobile ConvNet can be modeled as the sum of each $i$-th layer's 
runtime $R^{i}$, since the runtime of each operator
is independent of other operators~\cite{cai2017neuralpower,cai2018proxylessnas,wu2018fbnet}:
\begin{equation}
    \label{eq:runtime-network}
    R(\textbf{w} | \textbf{t}_{k}, \textbf{t}_{e}) = 
    \sum_{i} R^{i}(\textbf{w}^{i} | \textbf{t}^{i}_{k}, \textbf{t}^{i}_{e}, \textbf{t}^{i}_{se})
\end{equation}

For our approach, we adapt the per-layer runtime model as a function of the 
NAS-related decisions $\textbf{t}$.
We profile the target mobile platform (Pixel 1)
and we record the runtime for each candidate kernel operation per layer $i$, 
\textit{i.e.}, $R^{i}_{3 \times 3,3}$, $R^{i}_{3 \times 3,6}$, $R^{i}_{5 \times 5,3}$,
and $R^{i}_{5 \times 5,6}$. We denote the runtime of layer $i$ by following the notation in Equation~\ref{eq:proposed-form-3}. First, we express the runtime of each layer $i$ as a
function of the expansion ratio decision:
\begin{equation}
    \label{eq:runtime-layer-e}
    \begin{split}
    R^{i}_{e} & = \mathbbm{1}(\left\Vert 
    \textbf{w}_{k, 3} \right\Vert^2 > t_{e=3}) \cdot 
    ( R^{i}_{5 \times 5, 3} + \\ & \mathbbm{1}(\left\Vert 
    \textbf{w}_{k, 6 \setminus 3} \right\Vert^2 > t_{e=6}) 
    \cdot ( R^{i}_{5 \times 5, 6} - R^{i}_{5 \times 5, 3} ))
    \end{split}
\end{equation}
By incorporating the kernel size decision, the runtime based on the kernel $k$ and expansion ratio decision $e$ is:
\begin{equation}
    \label{eq:runtime-layer-k}
    \begin{split}
    R^{i}_{k,e} = &  \frac{R^{i}_{3 \times 3,6}}{R^{i}_{5 \times 5,6}} \cdot R^{i}_{e} + \\
    R^{i}_{e} & \cdot (1-\frac{R^{i}_{3 \times 3,6}}{R^{i}_{5 \times 5,6}}) \cdot 
    \mathbbm{1}(\left\Vert 
    \textbf{w}_{5 \times 5 \setminus 3 \times 3} \right\Vert^2 > t_{k=5})
    \end{split}
\end{equation}

Next, we capture the effect that the SE path has on the runtime. We denote the total runtime of the $i$-th MBConv layer with kernel size $k$,  expansion ratio $e$, and SE ratios $0.25$ or $0.5$ as $R^{i}_{k \times k,e, se=0.25}$ and $R^{i}_{k \times k,e, se=0.5}$, respectively. Similarly, we denote the runtime of the MBConv layer without a SE path as $R^{i}_{k \times k,e, se=0}$. For notation clarity, let us define the relative increase in runtime due to the addition of the SE path, compared to the runtime without the SE path, as scaling factor:
\begin{equation}
    \label{eq:runtime-scale}
    s^{i}_{k,e,se} = R^{i}_{k \times k,e, se} / R^{i}_{k \times k,e, se=0}
\end{equation}

Based on our runtime profiling (Section~\ref{sec:setup}), we make two observations: (i) due to the relatively smaller size of the \textit{squeeze} convolution compared to the $k \times k $ convolution of the main path, the difference in the relative runtime increase from using either SE ratios is negligible, \textit{i.e.}, $s^{i}_{k,e,0.25} \approx s^{i}_{k,e,0.5}$. Next, (ii) the relative ratio of the runtimes with and without the SE path differs based on the type of the main MBConv path. Thus, we express the overall runtime scaling as function of the kernel and the expansion ratio choices:
\begin{equation}
    \label{eq:runtime-scale-1}
    \begin{split}
    s^{i}_{k,e=6,0.25} & =  \mathbbm{1}(\left\Vert \textbf{w}_{5 \times 5 \setminus 3 \times 3} \right\Vert^2 > t_{k=5}) \cdot s^{i}_{k=5,e=6,0.25} + \\
             & (1-\mathbbm{1}(\left\Vert \textbf{w}_{5 \times 5 \setminus 3 \times 3} \right\Vert^2 > t_{k=5}) \cdot s^{i}_{k=3,e=6,0.25}
    \end{split}
\end{equation}
\begin{equation}
    \label{eq:runtime-scale-2}
    \begin{split}
    s^{i}_{k,e=3,0.25} & =  \mathbbm{1}(\left\Vert \textbf{w}_{5 \times 5 \setminus 3 \times 3} \right\Vert^2 > t_{k=5}) \cdot s^{i}_{k=5,e=3,0.25} + \\
             & (1-\mathbbm{1}(\left\Vert \textbf{w}_{5 \times 5 \setminus 3 \times 3} \right\Vert^2 > t_{k=5}) \cdot s^{i}_{k=3,e=3,0.25}
    \end{split}
\end{equation}
Hence, overall we have:
\begin{equation}
    \label{eq:runtime-layer}
    \begin{split}
    R^{i} = & \big(1 - \mathbbm{1}(\left\Vert \textbf{w}_{0.5 \setminus 0.25} \right\Vert^2 > t_{se=0.25})\big) \cdot R^{i}_{k,e} + \\
            & \mathbbm{1}(\left\Vert \textbf{w}_{0.5 \setminus 0.25} \right\Vert^2 > t_{se=0.25}) \cdot  \\
            & \Big\{ \mathbbm{1}(\left\Vert \textbf{w}_{k, 6 \setminus 3} \right\Vert^2 > t_{e=6}) \cdot s^{i}_{k,e=6,0.25} + \\
            & \big(1  -\mathbbm{1}(\left\Vert \textbf{w}_{k, 6 \setminus 3} \right\Vert^2 > t_{e=6})\big) \cdot s^{i}_{k,e=3,0.25} \Big\} \cdot R^{i}_{k,e}
    \end{split}
\end{equation}
As in Equation~\ref{eq:proposed-form-2}, we relax the indicator function to a sigmoid function $\sigma(\cdot)$ when computing gradients. By using this model, the runtime term in the loss function remains differentiable with respect to layer-wise NAS choices.

\begin{table*}[t!]
\caption{\textit{Single-Path} NAS achieves state-of-the-art image classification accuracy (\%) on ImageNet for similar on-device latency setting compared to previous NAS methods ($\sim 80 ms$ on Pixel 1), with up to $5,000 \times$ reduced search cost in terms of number of epochs.}
\begin{center}
\begin{tabular}{|c||c|c|c|c|}
\hline
Method\footnotemark & Top-1 Acc (\%) & Top-5 Acc (\%) & Runtime (ms) & Search  Cost (epochs) \\
\hline\hline
  
MobileNetV2~\cite{sandler2018mobilenetv2} & 72.00 & 91.00 & 75.00 & \multirow{3}{*}{-}  \\
MobileNetV2 (our impl.) & 73.59 & 91.41 & 73.57  &  \\
MobileNetV3~\cite{howard2019searching}  & 75.20 & -- & 78.00$\dag$ &  \\\hline

Random search     & 73.78 $\pm$ 0.85 & 91.42 $\pm$ 0.56 & 77.31 $\pm$ 0.9 ms & - \\\hline

MnasNet-B1~\cite{tan2018mnasnet} & 74.00 & 91.80 & 76.00 & \multirow{3}{*}{40,000}  \\
MnasNet-B1 (our impl.)  & 74.61 & 91.95 & 74.65 &   \\
MnasNet-A1~\cite{tan2018mnasnet}  & 75.20 &  92.50 & 78.00 &   \\\hline
MnasNet-B1 (92)~\cite{tan2018mnasnet}  & 74.79 &  92.05 & 92.00 &   \\\hline
ChamNet-B~\cite{dai2018chamnet}   & 73.80 & -- & -- & 240$\ddag$  \\\hline
ProxylessNAS-R~\cite{cai2018proxylessnas} & 74.60 & 92.20 & 78.00 & \multirow{2}{*}{200*}  \\
ProxylessNAS-R (our impl.)  & 74.65 & 92.18 & 77.48 &   \\\hline
FBNet-B~\cite{wu2018fbnet} & 74.1 & - & - & \multirow{2}{*}{90}  \\
FBNet-B (our impl.)  & 73.70 & 91.51 & 78.33 &   \\\hline
\hline 
\textbf{Single-Path} NAS (\textbf{proposed}) & \textbf{75.62} & \textbf{92.61} & 81.84 & \textbf{8} (\textbf{2.45 hours})  \\
\hline 
\end{tabular}
\end{center}

\label{tab:imagenet-sota}
\end{table*}

\section{Experimental Setup}
\label{sec:setup}

We use \textit{Single-Path NAS} to design ConvNets for image classification 
on ImageNet~\cite{deng2009imagenet}. We use Pixel 1 as the target 
mobile platform. The choice of this experimental setup is important,
since it allows for a representative comparison with prior 
hardware-efficient NAS methods that optimize for the same
Pixel 1 device around a target latency of 
$80ms$~\cite{cai2018proxylessnas,tan2018mnasnet}.

\textbf{Implementation and deployment}: We implement our NAS 
framework in TensorFlow (\texttt{TF} version 1.12).
During both search and training stages, we use TPUs 
(version 3)~\cite{jouppi2017datacenter}. To this end,  
we build on top of the \texttt{TPUEstimator} classes following the 
TPU-related documentation of the MnasNet 
repository~\cite{google2018mnasnet}. 
Last, all models (ours and prior work) are deployed with 
TensorFlow TFLite to the mobile device. On the device, we profile 
runtime using the Facebook AI Performance Evaluation 
Platform (\texttt{FAI-PEP})~\cite{facebook2018fai}
that supports profiling for \texttt{tflite} models with detailed 
per-layer runtime breakdown.

\begin{figure}[h!]
  \centering
  \includegraphics[width=.5\columnwidth]{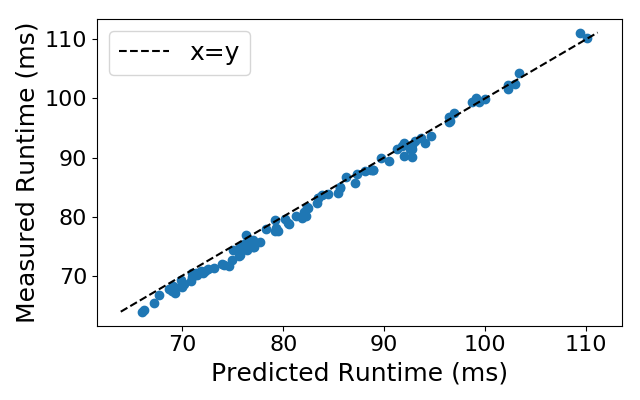}~
  \includegraphics[width=.5\columnwidth]{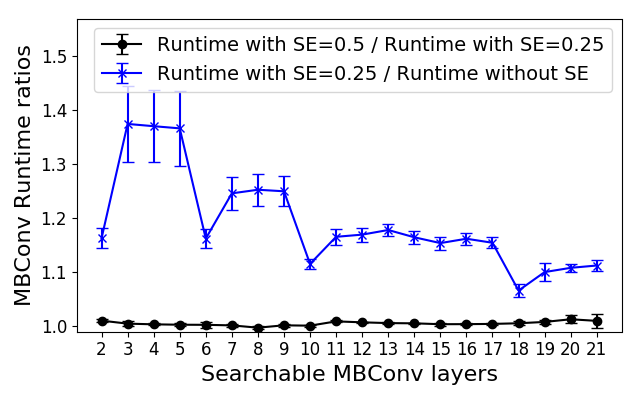}
  \caption{Runtime profiling: (Left) The runtime model (Equation~\ref{eq:runtime-network}) is accurate, with an average prediction error of $1.76\%$. (Right) Runtime results with SE ratios 0, 0.25, and 0.5 show that allowing for SE ratios larger than $0.25$ (\textit{i.e.}, 0.5 SE ratio) provides a better accuracy-runtime trade-off, since the \textit{squeeze} step is enhanced with more channels with negligible runtime overhead ($s^{i}_{k,e,0.25} \approx s^{i}_{k,e,0.5}$), especially for the deeper layers (MBConv 18-21).}
  \label{fig:runtime_lut}
\end{figure}

\textbf{Runtime model}: To train the inference runtime model, we record the runtime per layer (MBConv operations breakdown) by profiling ConvNets with all different MBConv types (Equations~\ref{eq:runtime-layer-e}-\ref{eq:runtime-layer}). To evaluate the runtime-prediction accuracy of the model, we generate 100 randomly designed ConvNets (with $se=0$) and we measure their runtime on the device. As illustrated in Figure~\ref{fig:runtime_lut} (left), our predictive model is accurate: the Root Mean Squared Error (RMSE) is $1.32ms$, which corresponds to an average $1.76\%$ prediction error. 

\textbf{Superkernels implementation}: We use \texttt{Keras} 
to implement our trainable ``superkernels.'' Specifically, we define a custom
\texttt{Keras}-based depthwise convolution kernel where the output is a function
of both the weights and the threshold-based decisions 
(Equations~\ref{eq:proposed-form-2}-\ref{eq:proposed-form-3}). Our custom 
layer also returns the effective runtime of the layer 
(Equations~\ref{eq:runtime-layer-e}-\ref{eq:runtime-layer}). 
We document our implementation in our project GitHub 
repository: \url{https://github.com/dstamoulis/single-path-nas},
with detailed steps on how to reproduce the results.

\section{State-of-the-art Runtime-Constrained ImageNet Classification}
\label{sec:results}

We apply our method to design ConvNets for the Pixel 1 phone with an overall target latency around $\sim 80ms$. We train the derived \textit{Single-Path} NAS model for 350 epochs, following the MnasNet training schedule~\cite{tan2018mnasnet}. We compare our method with mobile ConvNets designed by human experts and state-of-the-art NAS methods in Table~\ref{tab:imagenet-sota}, in terms of classification accuracy, search cost and hardware efficiency (inference latency on Pixel 1). To ensure a fair comparison, we retrain the baseline models following the same schedule (in fact, we find that the MnasNet-based training schedule improves the top1 accuracy compared to what is reported in several previous methods). Similarly, we profile the models on the same Pixel 1 device. For prior work that does not optimize for Pixel 1, we retrain and profile their model closest to the MnasNet baseline (\textit{e.g.}, the FBNet-B and ChamNet-B networks~\cite{dai2018chamnet,wu2018fbnet}, since the authors use these ConvNets to compare against the MnasNet model). Finally, we directly report the number of epochs reported per method, hence canceling out the effect of different hardware systems (GPU \textit{vs}. TPU hours).

\textbf{ImageNet classification}:
Table~\ref{tab:imagenet-sota} shows that our \textit{Single-Path} NAS achieves top-1 accuracy of $\textbf{75.62\%}$, which is the new state-of-the-art ImageNet accuracy among hardware-efficient NAS methods. More specifically, our method achieves \textbf{better} top-1 accuracy than ProxylessNAS by almost $1\%$, while maintaining on par target latency of $\sim 80ms$ on the same target platform. Overall, we note that \textit{Single-Path} NAS outperforms prior NAS methods in this mobile latency range~\cite{wu2018fbnet,tan2018mnasnet,dai2018chamnet}, as well as manually designed models (MobileNetV2~\cite{sandler2018mobilenetv2}) and ConvNets that combine both AutoML and manual-design expertise (MobileNetV3~\cite{howard2019searching}), \textit{e.g.}, better than MnasNet-A1 ($+0.42\%$), FBNet-B ($+1.52\%$), and MobileNetV3 ($+0.42\%$).

\textbf{Search cost}: \textit{Single-Path} NAS has orders of magnitude 
\textit{reduced} search cost compared to all previous hardware-efficient NAS methods.
Specifically, MnasNet reports that the controller uses 8k sampled models, each 
trained for 5 epochs, for a total of 40k train epochs. In turn, ChamNet
trains an accuracy predictor on 240 samples, which assuming an aggressively 
fast training schedule of five epochs per sample (same as in MnasNet),
corresponds to a total search cost of 1.2k epochs.
ProxylessNAS reports $200 \times$ search cost improvement over MnasNet,
hence the overall cost is the TPU-equivalent of 200 epochs. 
Finally, FBNet reports 90 epochs of training on a proxy dataset (10\%
of ImageNet). While the number of images per epoch is reduced, we found
that a TPU can accommodate a FBNet-like supermodel with maximum batch size
of 128, hence the number of steps per FBNet epoch are still $8 \times$
more compared to the steps per epoch in our method.

\begin{figure}[h!]
  \centering
  \includegraphics[width=.5\columnwidth]{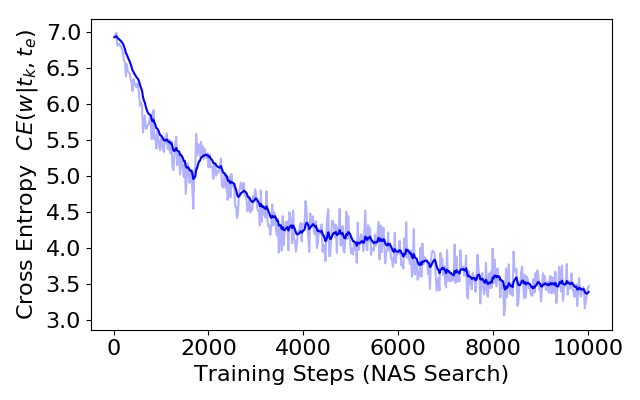}~
  \includegraphics[width=.5\columnwidth]{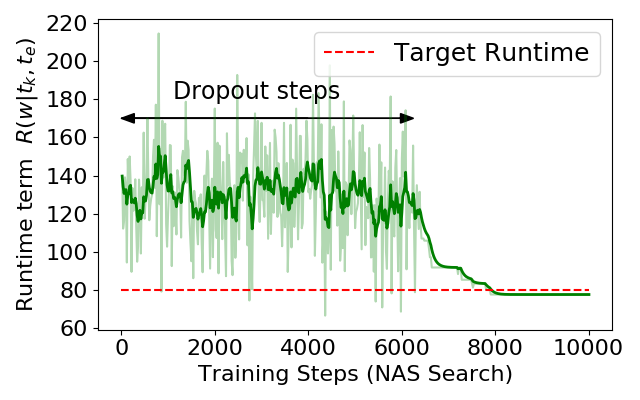}
  \caption{\textit{Single-Path NAS} search progress: Progress of both objective terms, 
  \textit{i.e.}, cross entropy $CE$ (left) and runtime $R$ (right) during NAS search.}
  \label{fig:progress}
\end{figure}

\begin{figure*}[t]
  \centering
  \includegraphics[width=0.9\textwidth]{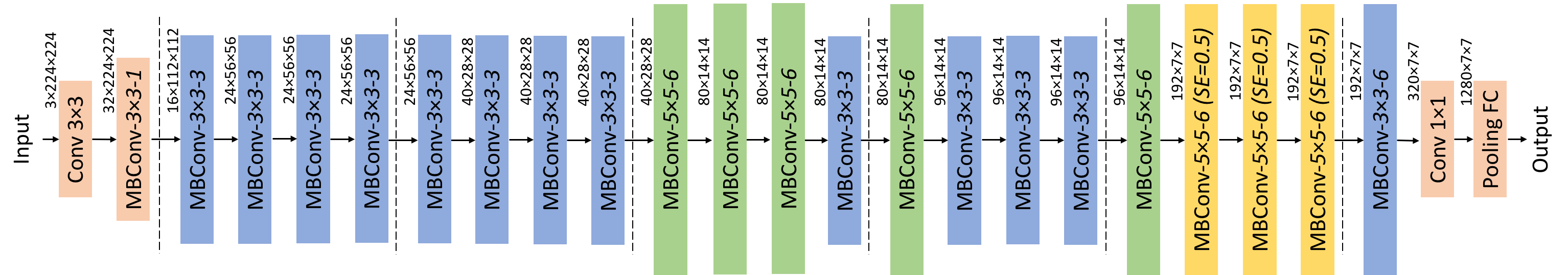}
  \caption{Hardware-efficient ConvNet found by \textit{Single-Path} NAS, with 
  top-1 accuracy of $\textbf{75.62\%}$ on ImageNet and inference time of $81.84 ms$ 
  on Pixel 1 phone. Compared to our previous NAS result without SE~\cite{stamoulis2019single},  some of the earlier $5 \times 5$ MBConvs have been replaced with smaller $3\times3-3$ MBConvs, and instead \textit{Single-Path NAS} selects SE paths with SE ratio of $se=0.5$ in the last layers. Overall, our NAS enhancement with fully searchable SE improves the accuracy-runtime trade-off of mobile ConvNets.} 
  \label{fig:spnet}
\end{figure*}

\footnotetext{Table~\ref{tab:imagenet-sota}: *The search cost
in epochs is estimated based on the claim~\cite{cai2018proxylessnas}
that ProxylessNAS is $200 \times$ faster than MnasNet. $\ddag$ChamNet does not detail
the model derived under runtime constraints~\cite{dai2018chamnet} 
so we cannot retrain or measure the latency. $\dag$~For MobileNetV3, we report the version that matches the MnasNet space backbone, since some additional manual enhancements in the network head are directly applicable to all other ConvNets considered.}

In comparison, \textit{Single-Path NAS} has a total cost of eight epochs, which is $\textbf{5,000} \times$ faster than MnasNet, $25 \times$ faster than ProxylessNAS, and $11 \times$ faster than FBNet. In particular, we use an aggressive training schedule similar to the few-epochs schedule used in MnasNet to train the individual ConvNet samples~\cite{tan2018mnasnet}. 
Overall, we visualize the search efficiency of our method in Figure~\ref{fig:progress}, where we show the progress of both $CE$ and $R$ terms of Equation~\ref{eq:sp-nas}. Earlier during our search (first six epochs), we employ \textit{dropout} across the different subsets of the kernel weights (Figure~\ref{fig:progress}, right). Dropout is a common technique in NAS methods to prevent the supernet from learning as an ensemble. Unlike prior art that employs this technique over the separate paths of the \textit{multi-path} supernet, we directly drop randomly the subsets of the superkernel in our \textit{single-path} search space. We search for $\sim 10k$ steps (8 epochs with a batch size of $1024$), which corresponds to total wall-clock time of \textbf{2.45 hours} on a TPUv3 (\textit{i.e.}, 24 TPU-hours).

\textbf{Enhancing accuracy-runtime trade-off}: Our derived ConvNet architecture is shown in Figure~\ref{fig:spnet}. Our goal is to understand the better accuracy-runtime trade-off  achieved by the searchable SE. To this end, a comparison against the earlier version of our work~\cite{stamoulis2019single} without SE can give insightful observations. In particular, we observe that, compared to the ConvNet previously derived in~\cite{stamoulis2019single}, some of the earlier MBConv types with either $5 \times 5$ kernels or expansion ration $6$, have been replaced with smaller $3\times3-3$ MBConvs, and instead the \textit{Single-Path NAS} flow selects SE paths with SE ratio of $se=0.5$ in the last few layers. Compared to the previous result without SE ($74.96\%$~\cite{stamoulis2019single}), we confirm that the use of SE improves the accuracy-runtime trade-off of mobile ConvNets, as attested by the top1 accuracy improvement while remaining around the same latency setting $\sim 80ms$.

In addition, to understand the NAS choices related to the SE paths in our ConvNet, we report the relative runtime increase per MBConv types for each layer in Figure~\ref{fig:runtime_lut} (right). We can make the following observations. First, we observe that the relative increase in the MBConv's runtime (scaling factor $s_{k,e,0.25}$ in Equation~\ref{eq:runtime-layer}) is closer to $1.0$ for the last 4 layers. This is to be expected, since the \textit{squeeze} $1\times 1$ convolution is performed on input feature maps with reduced spatial dimensions. Indeed, we observe that \textit{Single-Path NAS} appends SE paths in these last layers. Second, we notice that the difference in the relative runtime increase from using either SE ratios ($0.25$ or $0.5$) is negligible, \textit{i.e.}, $s^{i}_{k,e,0.25} \approx s^{i}_{k,e,0.5}$. This is important in the context of NAS, since prior work only searches over the binary decision of using $se=0.25$ or not, without searching for the $se$ value. Indeed, \textit{Single-Path NAS} selects $se=0.5$ for all the SE paths when included.

\textbf{Comparison with random search}: An increasing amount of recent methods appeal to the practicality of random search as a simple, parameter-free NAS alternative~\cite{xie2019exploring}. It is therefore important to have a comparison of our result against random search. Specifically, we randomly sample ten ConvNets with predicted runtime from $75ms$ to $80ms$ (simple sampling by rejection). The average accuracy and runtime of the random samples are reported in Table~\ref{tab:imagenet-sota}. We observe that, while random search does not outperform NAS methods, the overall accuracy is comparable to MobileNetV2. This result highlights that the effectiveness of NAS methods heavily relies upon the properties of the MobileNetV2-based design space. We provide an extensive analysis in Section~\ref{sec:hyper}, where we comprehensively study the variance in solutions from differentiable NAS and random search methods.

\begin{figure}[t]
    \centering
    \includegraphics[width=0.9\columnwidth]{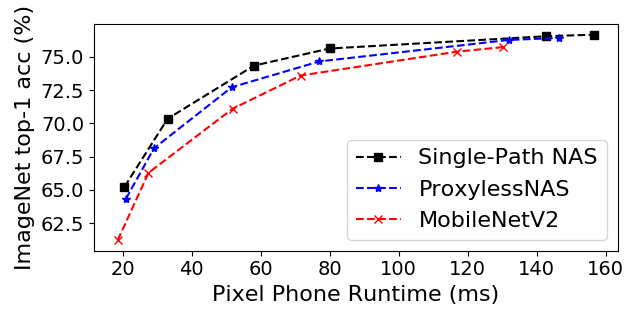}
    \vspace{-5pt}
    \caption{\textit{Single-Path} NAS outperforms MobileNetV2~\cite{sandler2018mobilenetv2} and ProxylessNAS~\cite{cai2018proxylessnas} across various channel size scales.}
    \label{fig:depth_mult_figure}
\end{figure}

\textbf{Channel scaling}: Next, we follow a typical analysis~\cite{cai2018proxylessnas,wu2018fbnet},
by rescaling the networks using a width multiplier~\cite{sandler2018mobilenetv2}.
As shown in Figure~\ref{fig:depth_mult_figure}, we observe 
that our model consistently outperforms prior methods under varying runtime settings. For instance, Single-Path NAS with $81.84ms$ is 1.44$\times$ faster
than the MobileNetV2 scaled model of similar accuracy.

\subsection{Ablation Study: Kernel-based Accuracy-Efficiency Trade-off}

\textit{Single-Path NAS} searches over subsets of the convolutional kernel weights. Hence, we conduct experiments to highlight how kernel-weight subsets can capture accuracy-efficiency trade-off effectively. To this end, we use the MobileNetV2 macro-architecture as a backbone (we maintain the  location of stride-2 layers as default). As two baseline networks,  we consider the default MobileNetV2 with MBConv-$3\times 3$-$6$ blocks  (\textit{i.e.}, $ \textbf{w}_{3 \times 3} $ kernels for all depthwise convolutions),  and a network with MBConv-$5\times 5$-$6$ blocks (\textit{i.e.}, $ \textbf{w}_{5 \times 5} $ kernels).

Next, to capture the subset-based training of weights during a \textit{Single-Path} 
NAS search, we consider a \textit{ConvNet} with MBConv-$5\times 5$-$6$ blocks, where we 
compute the loss of the model over two subsets, 
(i) the inner $ \textbf{w}_{3 \times 3} $ weights, and (ii) by also using the 
remaining $\textbf{w}_{5 \times 5 \setminus 3 \times 3} $ 
weights. For each loss computed over these subsets, we accumulate back-propagated 
gradients and update the respective weights, \textit{i.e.}, gradients are being 
applied separately to the inner and to the entire kernel per layer. We
follow training steps similar to the ``switchable'' 
training across channels as in~\cite{yu2018slimmable} (for the remaining 
training hyper-parameters we use the same setup as the default MnasNet).
As shown in Table~\ref{tab:levels}, we observe the final accuracy
across the kernel granularity, \textit{i.e.}, with the 
inner $ \textbf{w}_{3 \times 3} $ and the 
entire $ \textbf{w}_{5 \times 5} =  \textbf{w}_{3 \times 3}  + 
\textbf{w}_{5 \times 5 \setminus 3 \times 3} $ kernels, 
follows an accuracy change relative to ConvNets with
individually trained kernels. 

Such finding is significant in the context of NAS, since choosing over 
subsets of kernels can effectively capture the accuracy-runtime trade-offs 
similar to their individually trained counterparts. We therefore conjecture that
our efficient superkernel-based design search can be flexibly adapted 
and benefit the guided search space exploration in other RL-based
NAS methods. Beyond the NAS literature, our finding is closely 
related to Slimmable networks~\cite{yu2018slimmable} (SlimmableNets
limit however their analysis across the channel dimension).

\begin{table}[t!]
\caption{Searching across subsets of kernel weights: 
ConvNets with weight values trained over subsets 
of the kernels ($3\times3$ as subset of $5 \times 5$)
achieve performance (top-1 accuracy) similar to ConvNets 
with individually trained kernels.}
\centering
\label{my-label}
\scalebox{0.9}{
\begin{tabular}{|c||c|c|}
\hline
Method & Top-1  Acc (\%)  & Top-5  Acc (\%)   \\\hline\hline
  
Baseline ConvNet -                    & \multirow{2}{*}{73.59} & \multirow{2}{*}{91.41}   \\
$\textbf{w}_{3 \times 3} $ kernels &  &    \\\hline
Baseline ConvNet -  & \multirow{2}{*}{74.10} & \multirow{2}{*}{91.67}   \\ 
$\textbf{w}_{5 \times 5} $ kernels &  &    \\ \hline

\textit{Single-Path ConvNet} -  & \multirow{2}{*}{73.43} & \multirow{2}{*}{91.42}  \\
inference w/ $\textbf{w}_{3 \times 3} $ kernels  &  &   \\\hline
\textit{Single-Path ConvNet} - inference w/  & \multirow{2}{*}{73.86} & \multirow{2}{*}{91.72} \\
$\textbf{w}_{3 \times 3}  + \textbf{w}_{5 \times 5 \setminus 3 \times 3} $ kernels &  &  \\
\hline
\end{tabular}
}
\label{tab:levels}
\end{table}

\section{COCO Object Detection Performance}
\label{sec:results-coco}

In this Section, we assess the performance of \textit{Single-Path NAS} as a feature extractor for object detection application. In particular, we use our network as a drop-in replacement for the backbone featurizer in the Mask-RCNN model~\cite{he2017mask}, which is based on Feature Pyramid Network (FPN)~\cite{lin2017feature} as head and on our network as a backbone. Similarly, we train the model and we compare with other backbones networks, \textit{i.e.}, based on backbones from models designed from earlier mobile NAS methods. We train our model on the COCO dataset~\cite{lin2014microsoft}.

\begin{table}[h!]
\caption{COCO Object Detection Performance}
\centering
\label{my-label}
\scalebox{0.85}{
\begin{tabular}{|c||c|c|c|c|}
\hline
Method & $AP$ & $AP_S$ & $AP_M$ & $AP_L$  \\\hline\hline
MobileNet-V2 + Mask-RCNN & 30.47  & 16.49 &  32.33  & 41.14  \\\hline
MnasNet-B1 + Mask-RCNN & 32.47 &  17.74 &  34.45  & 43.88   \\\hline
ProxylessNAS + Mask-RCNN &  32.93  & 17.76  &  34.86  & 44.43   \\\hline
Single-Path NAS + Mask-RCNN (Proposed) &  \textbf{33.03} & \textbf{17.82} & \textbf{35.48} & \textbf{44.76}   \\
\hline
\end{tabular}
}
\label{tab:detection}
\end{table}

We use the open-source implementation of TPU-trained Mask-RCNN\footnote{\url{https://cloud.google.com/tpu/docs/tutorials/mask-rcnn}} for experiments. The models are trained on TPUs with batch size of 64. We train the different models on COCO \texttt{train2017} and we evaluate them on COCO \texttt{val2017}. Following typical the typical FPN flow~\cite{ghiasi2019fpn}, we attach the last feature extractor to the detection head. It is worth noticing that FPN is less hardware efficient compared to MobileNet-like alternatives such as SSDLite~\cite{sandler2018mobilenetv2}. Nonetheless, the focus of this analysis is to assess the various NAS designs are feature extractor while assuming the head design (ergo, the latency) fixed. Indeed, in Table~\ref{tab:detection} we observe that \textit{Single-Path NAS} outperforms other designs in terms of Average-precision (AP) and across all scales.

\section{Hyperparameter Optimization of Differentiable NAS}
\label{sec:hyper}

\subsection{Architecture Distribution in Differentiable NAS} 

While NAS literature has been traditionally driven by strong empirical results, the AutoML community has motivated studies to understand the properties of NAS solvers, their limitations, how and why they yield strong performance~\cite{bender2018understanding}. Hence, we find important to investigate the following questions: ``\textit{How do the different NAS formulations, \textit{e.g.}, the encoding of NAS choices across multiple paths or a single path, affect the differentiable NAS performance?}'' This is an important first step towards analyzing single-path formulations. 

Moreover, prior work on mobile NAS~\cite{wu2018fbnet,cai2018proxylessnas} lacks a detailed intra-level analysis on the statistics of differentiable methods, so a valid question to ask is: ``\textit{By how much does the quality of the ConvNet design vary across multiple runs of the same NAS search?}'' For instance, Stochastic NAS~\cite{xie2018snas} investigated the entropy of architecture distributions, but the analysis is limited to cell-based designs~\cite{liu2018darts} and does not consider mobile AutoML.

\begin{figure*}[t]
    \centering
    \includegraphics[width=0.95\textwidth]{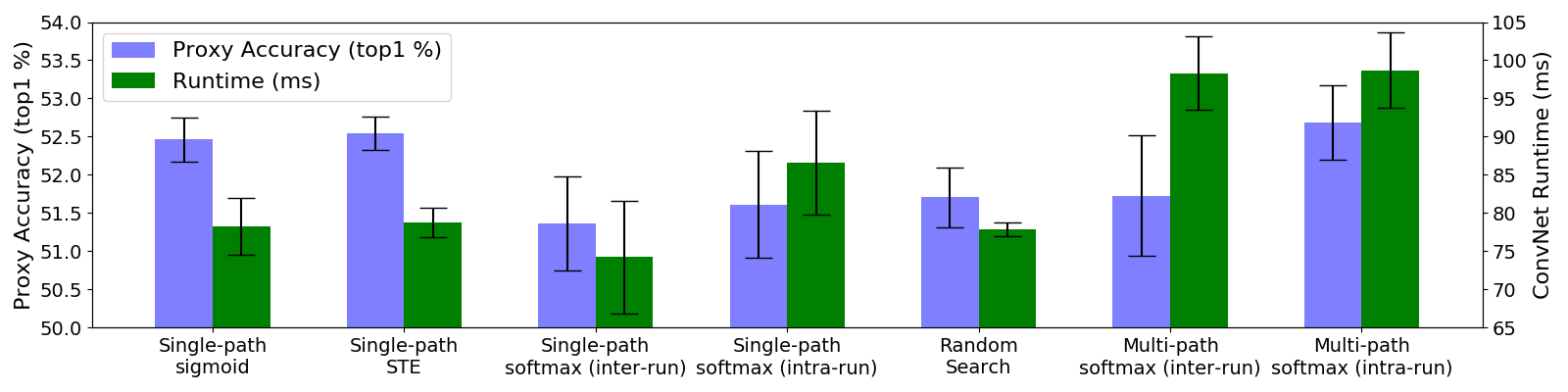}
    \vspace{-10pt}
    \caption{``\textit{How do the differentiable Mobile NAS formulation assumptions affect the overall performance (accuracy and runtime) of the AutoML-designed ConvNet?}'' Statistics (mean and variance) for the (proxy) accuracy (top 1$\%$) and the runtime of ConvNets designed via various formulations across 20 runs; for intra-run statistics, we pick the Pareto optimal ConvNet out of the 20 samples and we train another 20 ConvNets sampled from the softmax distribution.}
    \label{fig:entropy-analysis}
\end{figure*}

To quantitatively answer these two questions, we consider the following differentiable NAS formulations:

\noindent
\textbf{1. Multi-path with sigmoid}: This implementation solves the bilevel, multi-path formulation of Equation~\ref{eq:bilevel}. We implement a vanilla differentiable multi-path NAS solver~\cite{cai2018proxylessnas}. While our implementation replicates prior work's methodology~\cite{wu2018fbnet}, we adjust the solver to the aggressive few-epoch schedule used in~\cite{tan2018mnasnet}. This allows us to assess whether existing multi-path methods can reach a high-performing ConvNet within the same number of epochs as \textit{Single-Path NAS}.

Specifically, we set the number of total steps to eight epochs and we update the warm-up and learning rate schedules accordingly. We slim down the \textbf{multi-path supernet} by a width-multiplier factor of $0.5$ (recent NAS work also employs such search on a scaled-down model~\cite{tan2019efficientnet}). Similar to~\cite{wu2018fbnet}, we generate a proxy dataset (\textit{i.e.}, subset of ImageNet with 100 classes) to search on. We deploy our implementation (also available in our GitHub page) on cloud TPUs. 

Next, we investigate various \textbf{single-path}-based formulations:

\noindent
\textbf{2. Single-path with sigmoid}: this is the default implementation detailed in Section~\ref{sec:sp-methodology}. That is, during search (backpropagation over the supernet) we approximate the indicator functions (\textit{e.g.}, $\mathbbm{1}(\left\Vert \textbf{w}_{5 \times 5 \setminus 3 \times 3} \right\Vert^2 > t_{k=5})$) with sigmoid functions $\sigma ()$.

\noindent
\textbf{3. Single-path with STE}: during search we approximate the indicator functions (\textit{e.g.}, $\mathbbm{1}(\left\Vert \textbf{w}_{5 \times 5 \setminus 3 \times 3} \right\Vert^2 > t_{k=5})$) with the  straight-through estimator (STE)~\cite{yin2019understanding,bengio2013estimating}.

\noindent
\textbf{4. Single-path with softmax}: This implementation is a hybrid between the single-path encoding of the design space and the use of softmax, \textit{i.e.}, we encode the NAS choice of selecting across superkernel subsets using a softmax function parameterized by $\tau$, \textit{i.e.}, $\text{softmax} (\tau)$. For instance, we represent the kernel-level decision as:
\begin{equation}
    \label{eq:proposed-form-softmax-2}
    \textbf{w}_{k} = \frac{\text{exp}(\tau_{3\times 3})}{\sum_j \text{exp}(\tau_j)} \cdot \textbf{w}_{3 \times 3} + \frac{\text{exp}(\tau_{5\times 5})}{\sum_j \text{exp}(\tau_j)}  \cdot (\textbf{w}_{3 \times 3} + \textbf{w}_{5 \times 5 \setminus 3 \times 3})
\end{equation}
To update the kernel-level $\text{softmax}(\tau)$ choices, we formulate the \textit{Single-Path} search as a bilevel optimization problem $\underset{\tau}{\text{min }} \underset{\textbf{w}_{\tau}}{\text{min }} \mathcal{L}(\tau, \textbf{w}_{\tau})$, where the steps for updating the NAS $\tau$ parameters and the ConvNet weights occur interchangeably. 

\noindent
\textbf{5. Random search}: Parameter-free random search via constrained sampling. That is, we employ simple sampling by rejection, \textit{i.e.}, we keep the samples with runtimes within the range of interest $\sim 80ms$.

For all the aforementioned methods, we find the $\lambda$ value that achieves the desired accuracy trade-off $\sim 80ms$ (to tune $\lambda$, we use the hyperparameter-tuning scheduler presented in the next subsection~\ref{subsec:hyper}). We repeat the same NAS search experiment 20 times and we measure the mean and (\textbf{inter-}) variance across  the 20 runs for both objective terms, \textit{i.e.}, validation accuracy and runtime of the AutoML-designed ConvNet, denoted as \textit{inter-run}. In addition, to capture the (\textbf{intra-}) variance within a single search in softmax-based methods, we pick the best result among the 20 runs, and we train 20 new samples from the softmax distribution (in fact, similar selection is used in~\cite{wu2018fbnet} where 10 ConvNets are sampled and trained to pick the best). We denote the latter variant as \textit{intra-run}. We train each ConvNet for few epochs to obtain a representative proxy-accuracy value, following the aggressive training used in Mnasnet to study their RL method~\cite{tan2018mnasnet}. We summarize our results in Figure~\ref{fig:entropy-analysis}. 

\textbf{Comparison \textit{vs.} random search}: This result is particularly interesting, since there has been recent discussion within the NAS community on whether simple random search could find designs with performance comparable to those of more complex methods~\cite{li2019random}. Indeed, we observe that random search performs on par with multi-path cases, which confirms similar observations by recent work~\cite{xie2019exploring,cho2019one}. Nonetheless, it is important to note that random search is still inferior compared to \textit{Single-Path NAS} in terms of the (proxy) accuracy around the target latency range $\sim 80ms$. 

Furthermore, the nearly-zero search cost of random search is not necessarily representative: to avoid training all random, constraint-satisfying samples, an AutoML practitioner would employ an evaluation on the proxy task, by training each sample for few epochs and by picking the one with highest accuracy. Hence, the actual search cost for random search is not negligible. In fact, the low search cost of our method (8 epochs) is comparable to the number of training epochs during the aforementioned selection process. Given that \textit{Single-Path NAS} gives ConvNets with superior performance than random search at comparable cost, we argue that NAS remains a better AutoML options than random search methods.

\textbf{Softmax intra-run variance}: Next, we note the variance inherent to all the softmax-based cases.  That is, we observe that sampling the softmax of the best NAS search (selected from the 20 NAS repetitions) yields high-variance in terms of both accuracy and runtime. This finding confirms a recent analysis that shows the high entropy in the architecture distribution for cell-based multi-path designs~\cite{xie2018snas}.

\begin{figure*}[t!]
  \centering
  \includegraphics[width=.36\textwidth]{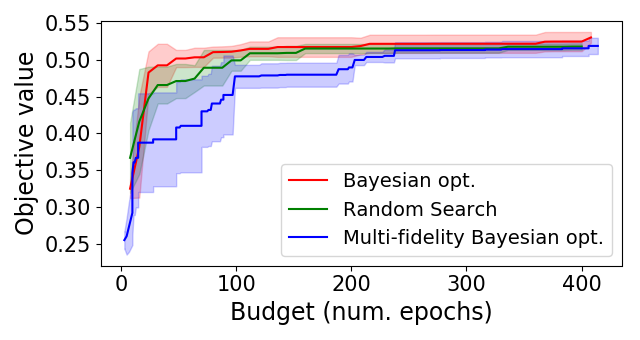}~
  \includegraphics[width=0.62\textwidth]{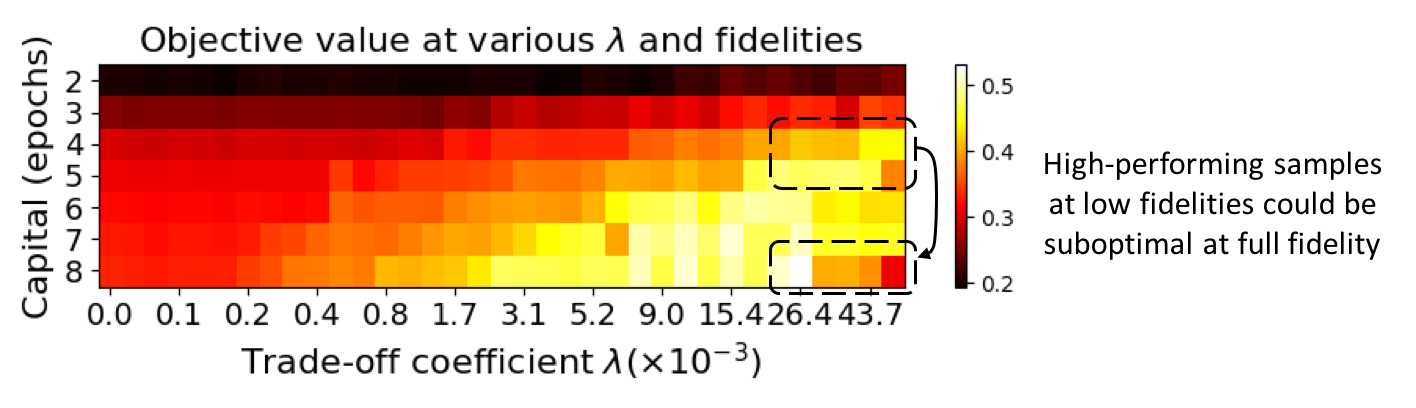}
  \vspace{-5pt}
  \caption{\textbf{Left}: Progress of various hyperparameter optimization solvers with respect to the overall reward. \textbf{Right}: Visualizing the objective value (Equation~\ref{eq:hyper-nas}) across multiple fidelities (y-axis) and hyperparameter values (x-axis) via grid search. Interestingly, low-cost function evaluations (middle, right) that reach the Pareto point around the target latency faster, tend to ``overshoot'' beyond this point towards over-constrained, suboptimal designs (bottom, right).}
  \label{fig:progress-dragonfly}
\end{figure*}

\textbf{Different single-path variants}: Moreover, we compare our original \textit{Single-Path NAS} (single-path sigmoid) method against its two variants (i) with STE and (ii) with softmax (inter-run). First, once again we note that the softmax version has higher variance compared to both the sigmoid and the STE versions. For the STE version, while the variance appears smaller than sigmoid, it is important to note that we had to repeat the process multiple times to reach 20 completed searches due to encountered numerical instability issues with STE (exploding gradients). A deeper study on the STE is an interesting direction for future NAS work, similar to recent STE analysis in the context of hardware-aware quantization~\cite{yin2019understanding}.

\textbf{Single-Path NAS \textit{vs.} prior work}: Last, we highlight the advantage of using our proposed method (single-path sigmoid) \textit{vs}. existing methods~\cite{wu2018fbnet,cai2018proxylessnas} (multi-path softmax, inter-run). We observe that the variance across different \textit{Single-Path NAS} runs is smaller than the variance of softmax-based methods (both inter- and intra-run).

Overall, we observe that multi-path softmax methods sample either low accuracy ConvNets (many layers skipped, which is another issue previously observed~\cite{xie2018snas}) or higher accuracy ones that violate the constraint. We hypothesize that the inferior solutions are due to the fact that the bilevel problem (Equation~\ref{eq:bilevel}) is an intrinsically more complex optimization problem to solve, as also discussed in~\cite{liu2018darts}. That is, it is difficult for the multi-path solver to reach a high quality solution within few epochs, while our proposed \textit{Single-Path NAS} for the same number of steps is as costly as training a compact model. 

Besides the optimization complexity, one would argue that the performance of multi-path methods is decided by several hyperparameters. Indeed, we extensively experimented with numerous settings by varying the number of epochs between the interleaved steps (NAS \textit{vs}. ConvNet weights updates), the learning rates for each update step, the batch size, the parameterization of the Gumbel-softmax~\cite{wu2018fbnet}, to name a few. Given that running each solver parameterization is expensive (hundreds of epochs), this highlights another limitation related to the tuning cost for all the hyperparameters involved, making our proposed method even more appealing to use.  In fact, in the next subsection, we aim to fully erase this engineering cost for the AutoML practitioner, by automatically tuning the hyperparameters of \textit{Single-Path NAS}.

\subsection{Hypertuning the NAS hyperparameterizer}
\label{subsec:hyper}

NAS methods approximate Pareto solutions by a customized weighted objective parameterized by a trade-off parameter $\lambda$~\cite{tan2018mnasnet}, but this value is manually picked. For instance, Mnasnet employs an empirical rule based on ``prior'' runtime-accuracy trade-off knowledge~\cite{tan2018mnasnet}, while FBNet~\cite{wu2018fbnet} and ProxylessNAS~\cite{cai2018proxylessnas} do not provide details on the $\lambda$ value used or how it was picked. Hence, we aim to answer the question: ``\textit{Instead of empirically tuning the trade-off hyperparameter, can we automatically find it for a target runtime given by the hardware engineers?}'' 

To this end, we formulate the tuning of $\lambda$ (Equation~\ref{eq:loss}) as a \textit{hyperparameter optimization} problem itself. Specifically, we solve for the $\lambda$ value that maximizes the validation accuracy around runtime target $R_T$. For a representative analysis, we use the weighted objective introduced in~\cite{tan2018mnasnet} that approximates Pareto optimal solutions, allowing our approach to traverse the Pareto front while solving for $\lambda$. Specifically, we write:
\begin{equation}
    \label{eq:hyper-nas}
    \begin{split}
    \underset{\lambda}{\text{max}} & \text{~~} Acc_{valid} (\lambda | \textbf{w}, \textbf{t}_{k}, \textbf{t}_{e}, \textbf{t}_{se}) \cdot 
    \bigg[ \frac{R (\lambda | \textbf{w}, \textbf{t}_{k}, \textbf{t}_{e}, \textbf{t}_{se})}{ R_T} \bigg]^{w} \\
    & \text{with~~~}  w = 
        \begin{cases}
            0,  & \text{if } R (\lambda | \textbf{w}, \textbf{t}_{k}, \textbf{t}_{e}, \textbf{t}_{se})
 \leq R_T\\
            -1,   & \text{otherwise}
        \end{cases}
    \end{split}
\end{equation}

We would like to stress here that each evaluation of Equation~\ref{eq:hyper-nas} corresponds to new NAS search. Therefore, solving this hyperparameter optimization problem would be impractical with previous NAS methods where each function evaluation would cost hundreds of hours. Instead, we exploit the efficiency of Single-Path NAS and we investigate various \textit{black-box} hyperparameter optimization techniques. Specifically, we consider the following methods: 

\noindent
\textbf{1. Bayesian optimization~\cite{shahriari2015taking}}: Vanilla Bayesian optimization, as implemented in the \texttt{Dragonfly} tool~\cite{kandasamy2019tuning}, available online\footnote{\url{https://github.com/dragonfly/dragonfly/}}. The method fits a Gaussian process (GP)~\cite{rasmussen2006gaussian} (probabilistic model) to the objective (Equation~\ref{eq:hyper-nas}) by points sampled across the hyperparameter $\lambda$.

\noindent
\textbf{2. Multi-fidelity optimization~\cite{kandasamy2017multi}}: Enchanced Bayesian optimization method where the GP fits both the hyperparameter space ($\lambda$ values) and the \textit{fidelity} (budget) space. The intuition is that low-fidelity evaluations could offer a good view of the function manifold at lower cost. We use discrete budget choices from two up to eight epochs (eight epochs is the default maximum in the vanilla case) as multiple fidelities. We use the multi-fidelity method from \texttt{Dragonfly}~\cite{kandasamy2019tuning} which, for each new sample to evaluate, suggests the $\lambda$ value and the sample budget (epochs). 

\noindent
\textbf{3. Random search~\cite{bergstra2012random}}: Parameter-free random search that randomly samples $\lambda$ values.

We extend our AutoML framework to support hyperparameter optimization. Our implementation automates the process of launching multiple (sequential or parallel) runs on cloud TPUs and calls the \textit{black-box optimization} solver that suggests the next sample to evaluate. Our goal is to find the trade-off $\lambda$ value that yields Pareto-optimal designs around the target runtime of $R_T = 80ms$. We run each solver for five runs with a total budget of 400 epochs and we track the current-best objective value. In Figure~\ref{fig:progress-dragonfly} (left), we report the objective value per hyperparameter optimization method, where we plot the average-best and the variance across the five runs.

\textbf{Vanilla \textit{vs}. multi-fidelity Bayesian optimization}: We observe that vanilla Bayesian optimization outperforms the multi-fidelity counterpart by reaching the near-optimal region faster and by converging to final solutions with higher reward. This is an interesting finding, since prior work shows that, for other hyperparameter settings (\textit{e.g}., learning rate) multi-fidelity enhances the optimization process~\cite{kandasamy2017multi}.

To fully investigate why this occurs, we employ a grid search across the budget epochs (from two to eight) and different $\lambda$ values, and we plot the objective value (Equation~\ref{eq:hyper-nas}) of each grid point in Figure~\ref{fig:progress-dragonfly} (right). Indeed, we can observe that the main assumption that ``low-cost samples give a representative view of the space''~\cite{kandasamy2017multi} does not fully hold. As highlighted in the Figure, we observe that initially promising $\lambda$ values (brighter objective values obtained after four or five epochs, middle right) become suboptimal (darker, bottom right). 

From a NAS design standpoint, the larger values $\lambda$ penalize the runtime term more so they approach the Pareto point around the target latency faster, but they tend to ``overshoot'' beyond this point towards over-constrained designs. We find this result interesting, since we postulate that other \textit{black-box} optimization techniques that rely on low-cost (early) approximation (\textit{e.g.}, Hyperband~\cite{li2016hyperband}) would encounter the same issue. Studying this hyperparameter optimization problem is an interesting research direction currently under-explored, so we aim to delve into this problem in future work.

\textbf{Comparison \textit{vs.} random search}: We find that random search, while never outperforming the Bayesian optimization result, has a relatively good performance at tuning $\lambda$. Interestingly, recent work shares similar observation when tuning NAS scaling hyperparameters via grid search~\cite{tan2019efficientnet}. We hope that our analysis would further foster exploration towards this direction, and we open-source our hyperparameter optimization scheduler as part of our framework. We aim to extend support for other methods, such as bandit-based optimization~\cite{li2016hyperband}.

\section{Conclusion}

In this paper, we proposed \textit{Single-Path NAS}, a NAS method that reduces the search cost for designing hardware-efficient ConvNets to \textbf{less than 3 hours}. The key idea is to revisit the one-shot supernet design space with a novel \textit{single-path} view, by formulating the NAS problem as \textit{finding which subset of kernel weights to use} in each ConvNet layer. We enhanced the accuracy-runtime trade-off in differentiable NAS by treating the Squeeze-and-Excitation path as a fully searchable operation with our \textit{single-path} encoding. \textit{Single-Path NAS} achieved $75.62\%$ top-1 accuracy on ImageNet, which is state-of-the-art accuracy compared to NAS methods around similar latency setting ($\sim 80ms$). More importantly, we reduced the NAS search cost down to only 8 epochs (24 TPU-hours), which is up to \textbf{5,000$\times$ faster} compared to prior work. 

Moreover, we exploited the efficiency of our method to answer questions related to the effectiveness of differentiable NAS. In particular, we studied how different NAS formulation choices affect the performance of the designed ConvNets. Last, we explored whether we can automatically find the NAS hyperparameters that yield the desired accuracy-runtime trade-off, by formulating the tuning of the NAS solver as a hyperparameter optimization problem itself.



\section*{Acknowledgment}
This research was supported in part by National Science Foundation CSR Grant No. 1815780 and National Science Foundation CCF Grant No. 1815899. Dimitrios Stamoulis also acknowledges support from the Qualcomm Innovation Fellowship (QIF) 2018 and the TensorFlow Research Cloud (TFRC) programs.

\ifCLASSOPTIONcaptionsoff
  \newpage
\fi




\bibliographystyle{IEEEtran}
\bibliography{IEEEabrv,./singlepathnas}

\end{document}